\documentclass{article} % For LaTeX2e

\usepackage[utf8]{inputenc}

\usepackage{url}
\usepackage{multirow}
\usepackage{lipsum}
\usepackage{amsthm}
\usepackage{mathtools}% Loads amsmathb
\usepackage{xfrac}
\usepackage[export]{adjustbox}
\usepackage{longtable}
\usepackage{booktabs}
\usepackage{colortbl}
\usepackage{xcolor}
\usepackage{graphicx}
\usepackage{subcaption}
\usepackage{verbatim}
\usepackage{placeins}
\usepackage{hyperref}
\hypersetup{
    colorlinks=true,
    urlcolor=[rgb]{0.188, 0.498, 0.886},
    linkcolor=[rgb]{0.188, 0.498, 0.886},    
    filecolor=magenta,      
    citecolor=[rgb]{0.188, 0.498, 0.886},
}
\usepackage{fancyvrb}
\usepackage{fixltx2e}
\usepackage{bera}
\usepackage{pdfpages}

\usepackage{pgf-umlsd}
\usepackage[draft=true]{minted}
\newcommand\aliasgreen[1]{\textcolor[HTML]{529E88}{#1}}

\usepackage{algorithm} % algorithm style specific (testing)
\usepackage{program} % algorithm style specific (testing)
\usepackage{algorithmic}%testing
\usepackage{listings}
\usepackage{geometry}

\usepackage{caption,setspace}
\DeclareCaptionFormat{listing}{\rule{\dimexpr0.9\columnwidth+17pt\relax}{0.4pt}\par\vskip1pt#1#2#3}
\newcounter{code}



\DeclareMathVersion{sans}
\SetSymbolFont{operators}{sans}{OT1}{cmbr}{m}{n}
\SetSymbolFont{letters}  {sans}{OML}{cmbrm}{m}{it}
\SetSymbolFont{symbols}  {sans}{OMS}{cmbrs}{m}{n}

\lstnewenvironment{sflisting}[1][]
  {\lstset{#1}\mathversion{sans}}{}

\usepackage[normalem]{ulem}

\definecolor{grayalias}{HTML}{3F4444}
\definecolor{bluealias}{HTML}{307FE2}
\usepackage{authblk}

\title{\aliasgreen{\bf DevSecOps} in Robotics}
\author[1,2]{Víctor Mayoral-Vilches}
\author[2]{Nuria García-Maestro\\}
\author[2]{McKenna Towers}
\author[2]{Endika Gil-Uriarte}

\affil[1]{
    {\small System Security (SYSSEC) group from Universität Klagenfurt, Austria.
        {\tt\footnotesize v1mayoralv@edu.aau.at}}
}

\affil[2]{
     {\small Alias Robotics, Vitoria-Gasteiz, Álava, Spain }}

\begin{document}
%\includepdf[pages=-, fitpaper]{SecDevOps_cover.pdf}

\maketitle

%===============================================================================
\vspace{-1em}
% 	Abstracts should be a single paragraph, between 4--6 sentences long, ideally.
\begin{abstract}

Quality in software is often understood as "execution according to design purpose" whereas security means that "software will not put data or computing systems at risk of unauthorized access." There seems to be a connection between these two aspects but, how do we integrate both of them in the robotics development cycle? In this article we introduce DevSecOps in Robotics, a set of best practices designed to help roboticists implant security deep in the heart of their development and operations processes. First, we briefly describe DevOps, introduce the value added with DevSecOps and describe and illustrate how these practices may be implemented in the robotics field. We finalize with a discussion on the relationship between security, quality and safety, open problems and future research questions.

\end{abstract}

%===============================================================================

\section{Introduction}

\begin{figure}[h!]
    \includegraphics[width=0.8\textwidth]{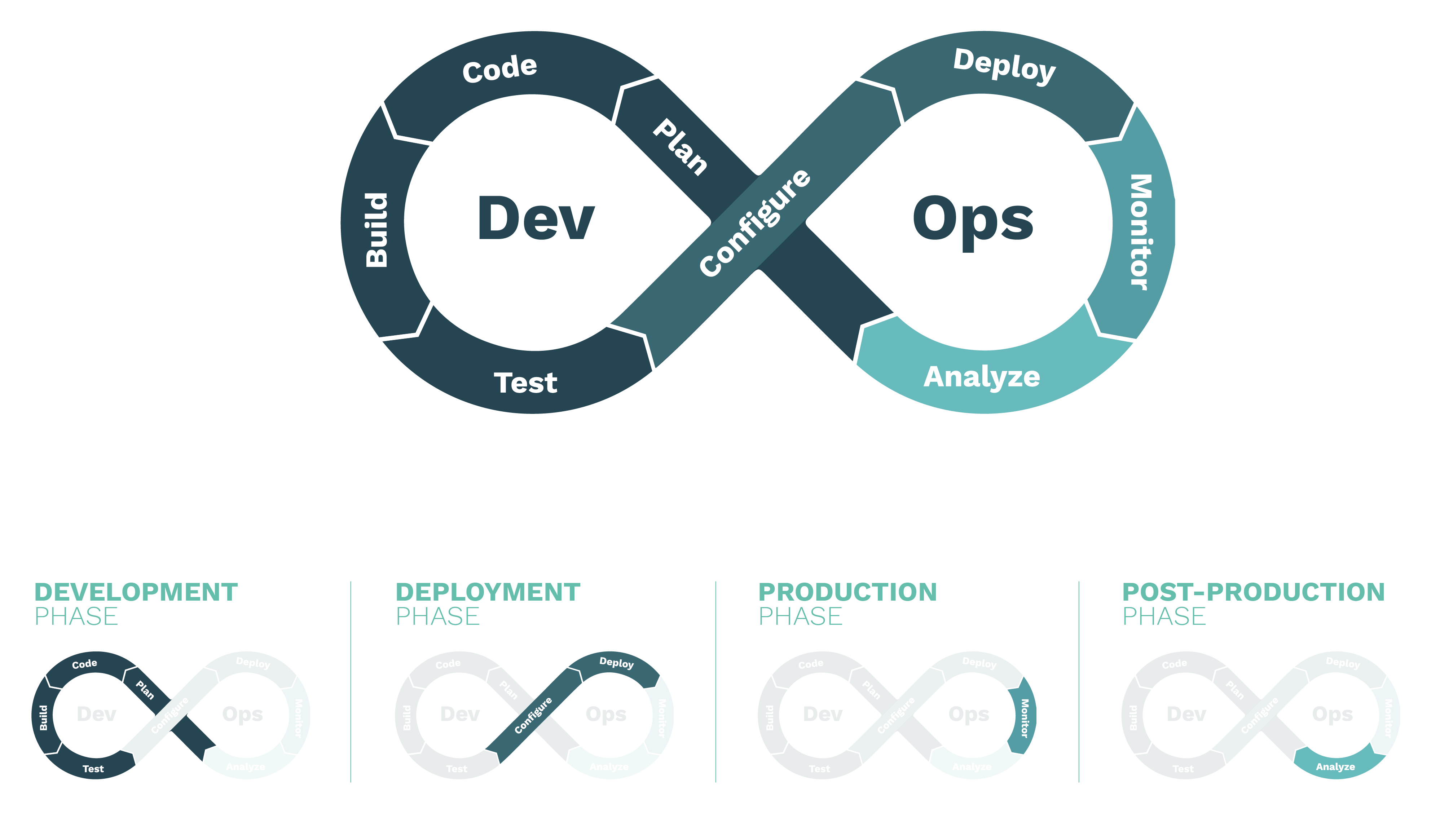}
    \centering
    \caption{The DevOps cycle and its four phases: a) development, b) deployment, c) production and d) post-production}
    \label{fig:devops_cycle}
\end{figure}

Where waterfall development once dictated the process of "handing off" from developers to Quality Assurance (QA) to operations, these responsibilities now have blurred borders with most engineers’ skill sets spanning multiple disciplines. Enter \textbf{DevOps}, a set of practices that combine software development (Dev) with Information Technology operations (Ops). They aim to shorten the development life cycle and provide continuous delivery while ensuring the quality of the software (Quality Assurance or QA)\cite{ebert2016devops}. Most sources \cite{devopsorigin1, devopsorigin2} agree that this idea began around 2008, with a discussion between Patrick Debois and Andrew Clay Shafer, concerning the concept of agile infrastructure. According to Roche \cite{roche2013adopting}, no standard definition exists for DevOps. What seems clear is that the design phase has been replaced by a less cyclic and more integrated flow evolution from development to demonstrations. \\
\newline
In Robotics, as in many other areas closely connected to Computer Science, there is a set of adopted common practices, including DevOps. Figure \ref{fig:devops_cycle} pictures the DevOps cycle most roboticists use and split into 4 phases: Development, Deployment, Production and Post-Production. In the DevOps philosophy, these phases are all connected in a continuous, never-ending, loop. The waterfall is now a river; development and operations flow together but, where is security in such a landscape? As pointed out by the UK's National Cyber Security Center \cite{ncsc_uk}, having a secure approach to development has never been so critical, when it comes to cyberphysical systems and, of course, robotics. The way roboticists build software and systems is rapidly evolving, becoming more and more automated and integrated. Many roboticists today define, prototype and develop an entire robotics system architecture in simulation and tie it to tooling which will automate both testing and deployment.\\
\newline
Over the last couple of years we have seen big changes on the arrival of cloud services to robotics and to 'robot infrastructure as a service' including simulation, fleet management systems, or teleoperation capabilities among others. The promise is that robots of almost any size and complexity can be called into virtual life, changed, or terminated without leaving the desktop. On top of these new capabilities, a process of quick and regular deployments has evolved. Often referred to as Continuous Delivery, this iterative approach is powerful, flexible and efficient, but these strengths bring new sets of risks which your security practices must address. To do so, roboticists will need to consider security as a primary concern throughout your development and deployment processes. \\
\newline
To the best of our knowledge, while there is some prior work studying DevOps in the embedded systems field \cite{lwakatare2016towards}, there is no literature available to date formalizing the adoption of DevSecOps in the robotics domain. In this article we introduce DevSecOps in Robotics as a set of best practices designed to help roboticists implant security deep in the heart of their development and operations processes. The content provided below is not an in-depth guidance on how to avoid implementation vulnerabilities in the code you write, this topic is covered by other pieces of research and we kindly refer the reader to \cite{graff2003secure, seacord2008cert, jones2004secure, seacord2005secure}. The principles provided below are intended to help secure the entire process of software development in robotics, from establishing a security-friendly culture in your organisation through implementation and ongoing management. It must be noted, however, that \underline{using these principles do not guarantee a secure final robot}, but should help you gain confidence and insights that the code you deploy is built with a security mindset.

%\todo{Speak about Quality, security and the origins of DevOps versus the waterfall model from recent article. Consider moving here the devops tihng.}

\section{From DevOps to DevSecOps}

\begin{figure}[h!]
    \includegraphics[width=0.9\textwidth]{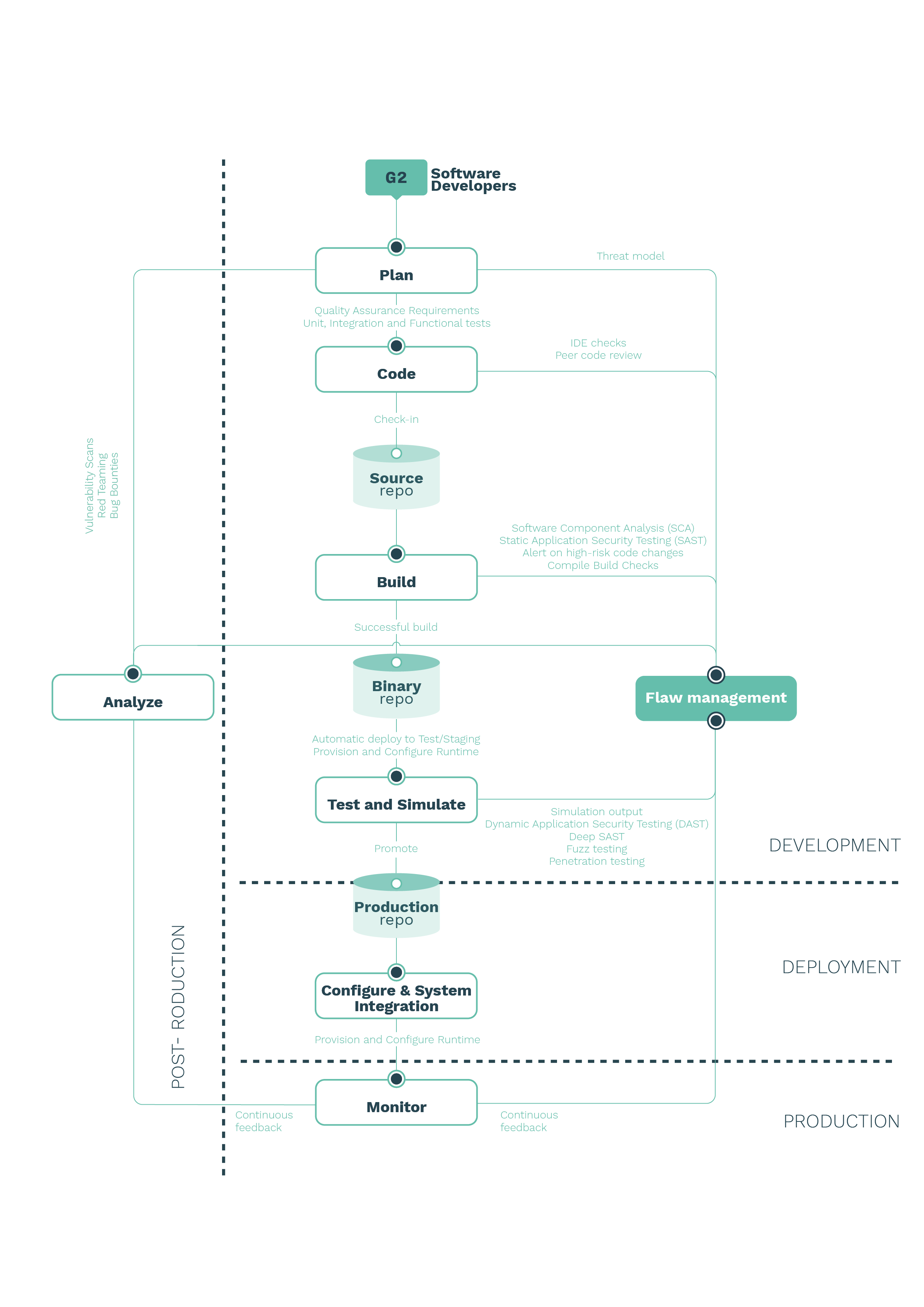}
    \centering
    \caption{A flowchart of DevSecOps for robotics}
    \label{fig:devsecops_diagram}
\end{figure}

Roche \cite{roche2013adopting} argues that developers work mostly on code while operations people work mostly with systems. Security, well identified as a continuous process, needs to be applied individually to each one of these phases and then holistically to the complete system. Security is a challenge and with each interface, the complexity increases.\\
\newline
DevOps is presented with a mix of the two skill sets mentioned above, developers and operations people. When adding security to the DevOps tuple we dive into DevSecOps. Often times called "Rugged DevOps" or “security at speed”,  \textbf{DevSecOps} is a set of best practices designed to help organizations implant security deep in their DevOps development and operations processes \cite{bird2016devopssec, mohan2016secdevops}. DevSecOps, also known as SecDevOps and DevOpsSec, seeks to embed security inside the development process as deeply as DevOps does with operations. \\
\newline
Turning again into robotics, security --identified as a process that needs to be continuously assessed both in hardware and software-- is a highly time and resource-consuming task. As was proven before in Computer Science, including security --across the development, deployment, production, and post-production phases-- results in a more secure output. Figure \ref{fig:devsecops_diagram} depicts our view on how DevSecOps could be implemented in robotics. We list phases, tasks, and activities while providing descriptions to further clarify the effort required.

\subsection*{Development \emph{phase}}
\subsubsection*{Plan \emph{task}}
\begin{itemize}
    \item \aliasgreen{\bf Threat modeling}: Threat modeling is the use of abstractions to aid in thinking about risks. A threat model identifies security threats that apply to the robot and/or its components (both software and hardware) while providing means to address or mitigate them in the context of a use case and ideally, during development.
    
    \item \aliasgreen{\bf Quality Assurance requirements}: Quality Assurance (QA) requirements imply capturing the quality metrics and aspects that will be later used to measure a code’s compliance. The requirements captured in this activity are later translated into actionable tests.

    \item \aliasgreen{\bf Unit, integration and functional tests}: Unit testing is the practice of testing small pieces of code, typically individual functions, alone and in an isolated manner. Integration testing involves checking whether different modules perform when combined together as a group.  Finally, functional testing verifies the functionality of a given software composition, that is to confirm that the corresponding source code provides the expected behavior. In line with Test-Driven Development and other groups' practices \cite{apexaisafealgs}, it is encouraged to write tests before coding. Particularly, we encourage to a) write the general code skeleton, classes, and public methods and to b) write the test cases for the logic to implement\footnote{Note that these tests should fail at this point}.
\end{itemize}

\subsubsection*{Code \emph{task}}
\begin{itemize}
    \item \aliasgreen{\bf IDE checks}: Many vulnerabilities in products and systems could be avoided if better, more secure coding practices were in place. A number  of  Integrated  Development  Environment  (IDE) plugins exist, which help developers check for security flaws while they code \cite{basetposter}.

    \item \aliasgreen{\bf Peer code review}: Peer Code Review, is the act of consciously and systematically analyzing fellow programmers’ code to check each other’s mistakes. Bosu et al. produced two interesting studies \cite{bosu2013peer, bosu2014characteristics} that researched vulnerabilities in code-while-in-development where they concluded that: a) the most experienced contributors authored the majority of the Vulnerable Code Changes (VCCs), b) Less experienced authors wrote fewer VCCs, but their code changes were 1.5 to 8 times more likely to be vulnerable, and c) employees of organizations sponsoring OSS (Open Source Software) projects were more likely to write VCCs. Coherently, peer code review is identified as an effective improvement practice for reducing security vulnerabilities. 
\end{itemize}

\subsubsection*{Build \emph{task}}
\begin{itemize}
    \item \aliasgreen{\bf Software Component Analysis (SCA)}: Software Component (or Composition) Analysis (SCA) is the process of identifying potential areas of risk from the use of third-party and open-source software components. In robotics, this could be very easily extended to hardware as well. SCA generally involves using a variety of different tools.
    \item \aliasgreen{\bf Static Application Security Testing (SAST)}: Static Application Security Testing, or static analysis, is the study of software without executing it and by simply looking at its structure and content. SAST is a testing methodology that analyzes source code to find security vulnerabilities.
    \item \aliasgreen{\bf Alert on high-risk code changes}: As a general recommendation, when the delta of code changes is above a certain pre-defined boundary, alerts should be generated.
    \item \aliasgreen{\bf Build (compilation) checks}: CI systems generally involve one or more building process. Compilation output can deliver valuable input and should be accounted for. It is good practice to monitor a compiler’s output and file tickets for review in the warnings that remain. This can be easily automated and tickets can be filed to a centralized flaw management system like the Robot Vulnerability Database (RVD) \cite{vilches2019introducing}. An open prototype of such a flaw management system, integrated with Github, is available at \url{https://github.com/aliasrobotics/RVD} including the necessary  tooling to manage it.
\end{itemize}

\subsubsection*{Test and simulate \emph{task}}
\begin{itemize}
    \item \aliasgreen{\bf Dynamic Application Security Testing (DAST)}:  Dynamic Application Security Testing (DAST) is a practice to detect conditions indicative of a security vulnerability in an application or software in its running state. The software is tested from the outside, that is, no access to the source code is assumed to be provided and thereby falls into the black-box security testing methodologies (against SAST, which requires access to the source code and is thereby considered a white-box security testing methodology). 
    \item \aliasgreen{\bf Deep Static Application Security Testing (Deep SAST)}: Past studies have shown \cite{barstad2014predicting, russell2018automated, gupta2018intelligent, pradel2018deepbugs} that it is possible to predict both code quality and security flaws using machine learning approaches. Deep Static Application Security Testing (Deep SAST) leverages on the the effectiveness of using machine learning-based static analysis to predict security flaws.
    \item \aliasgreen{\bf Fuzz testing}: Fuzz testing or fuzzing, is a semi-automated testing technique for software that consists of injecting data in search for errors. It involves providing invalid, unexpected, or random data as input. The software is then monitored for exceptions such as crashes, failing code assertions, or potential memory leaks. Similar to DAST, fuzzing is a black-box security testing methodology (no access to the source code).
    \item \aliasgreen{\bf Penetration Testing (PT or pentesting)}: Penetration testing or pentesting, is the practice of simulating a cyber attack against a given system (with a well defined scope) to check for exploitable vulnerabilities and determining its severity. Pentesting aims to find as many vulnerabilities and configuration issues as possible, in the time allotted, to then try exploit those to determine the risk of the vulnerability. A pentesting activity can be conducted as a white-box activity (complete access to source code), black-box (no access to sources or prior knowledge of the system) or gray-box (limited knowledge of the system available). Some penetration testing schemes are applicable to robotics, such as Robot Security Framework \cite{RSF}.
\end{itemize}

\subsubsection*{Flaw management \emph{task}}
\begin{itemize}
    \item \aliasgreen{\bf Flaw management activity}: Formally \cite{foreman2019vulnerability}, flaw management is the cyclical practice of identifying, classifying, re-mediating, and mitigating flaws. Flaw management is at the center of security activity. It refers to the process of how security researchers, suppliers, and end user organizations manage bugs and vulnerabilities, making systems more secure. \\
    \newline
    Flaw management requires a holistic, risk-based approach to security and combines the input received from other phases and tasks in the security process (e.g. SAST, DAST, build warnings, pentesting, fuzzing, etc.). The amount of flaws is often bigger than the engineering resources to manage them, thereby flaws should be prioritized. This is often done using a preliminary risk assessment, which is typically performed by using severity scoring mechanisms such as CVSS \cite{mell2006common} or RVSS \cite{vilches2018towards}, the latter more specific for robotics and accounting for safety hazards. 
\end{itemize}

\subsection*{Deployment \emph{phase}}
\begin{itemize}
    \item \aliasgreen{\bf Configuration and system integration task}: Deals with system integration and configuration of all parts previously developed and conforming the robotic system. It is one of the most sensitive tasks from a security perspective. It is critical to define secure deployment mechanisms and employ strong systematic configuration practices to ensure no flaws are presented during this task.
\end{itemize}

\subsection*{Production \emph{phase}}
\begin{itemize}
    \item \aliasgreen{\bf Monitor}: This task deals with the continuous monitoring of the system and should oversee the systems' performance, errors, and other relevant information. A common good practice is to employ elements that record all activity, securely, for potential future forensics research.
\end{itemize}

\subsection*{Post-production \emph{phase}}
\subsubsection*{Analyze \emph{task}}
\begin{itemize}
    \item \aliasgreen{\bf Vulnerability Scans}: Vulnerability scans provide results on new vulnerabilities affecting your system and provides solid data on their levels of severity. This data should be fed into the Flaw Management system and prioritized based on common scoring mechanisms.
    \item \aliasgreen{\bf Red Teaming}: Red teaming is a full-scope, holistic, multi-layered, and targeted (specific goals) attack simulation designed to measure how well a company’s systems, people, networks, and physical security controls can withstand an attack. Opposed to Penetration Testing, a red teaming activity does not seek to find as many vulnerabilities as possible to risk-assess them, instead it has a specific goal to look for vulnerabilities that will maximize damage and meet the goal. The ultimate objective of a red teaming activity is to test  organization/system detection and response capabilities in production. 
    \item \aliasgreen{\bf Bug Bounties}: A Bug Bounty program is a coordinated and well scoped effort to encourage security research in a particular system rewarding (with bounties) those researchers who help find bugs. Bug Bounty programs utilize a pay for results model, favour competition and often leverage the crowd sourced model. Bug bounty programs can be self hosted or coordinated by third parties. There are several online providers that offer support organizing the programs. When compared to other actions such as red teaming, bug bounty programs are generally result-oriented and add more diversity on the testers side (crowd-sourcing). On the down side, bug bounty programs do not generally have highly skilled security researchers and they require extensive program management (either on you or hired).
\end{itemize}

\begin{figure}[h!]
    \includegraphics[width=0.8\textwidth]{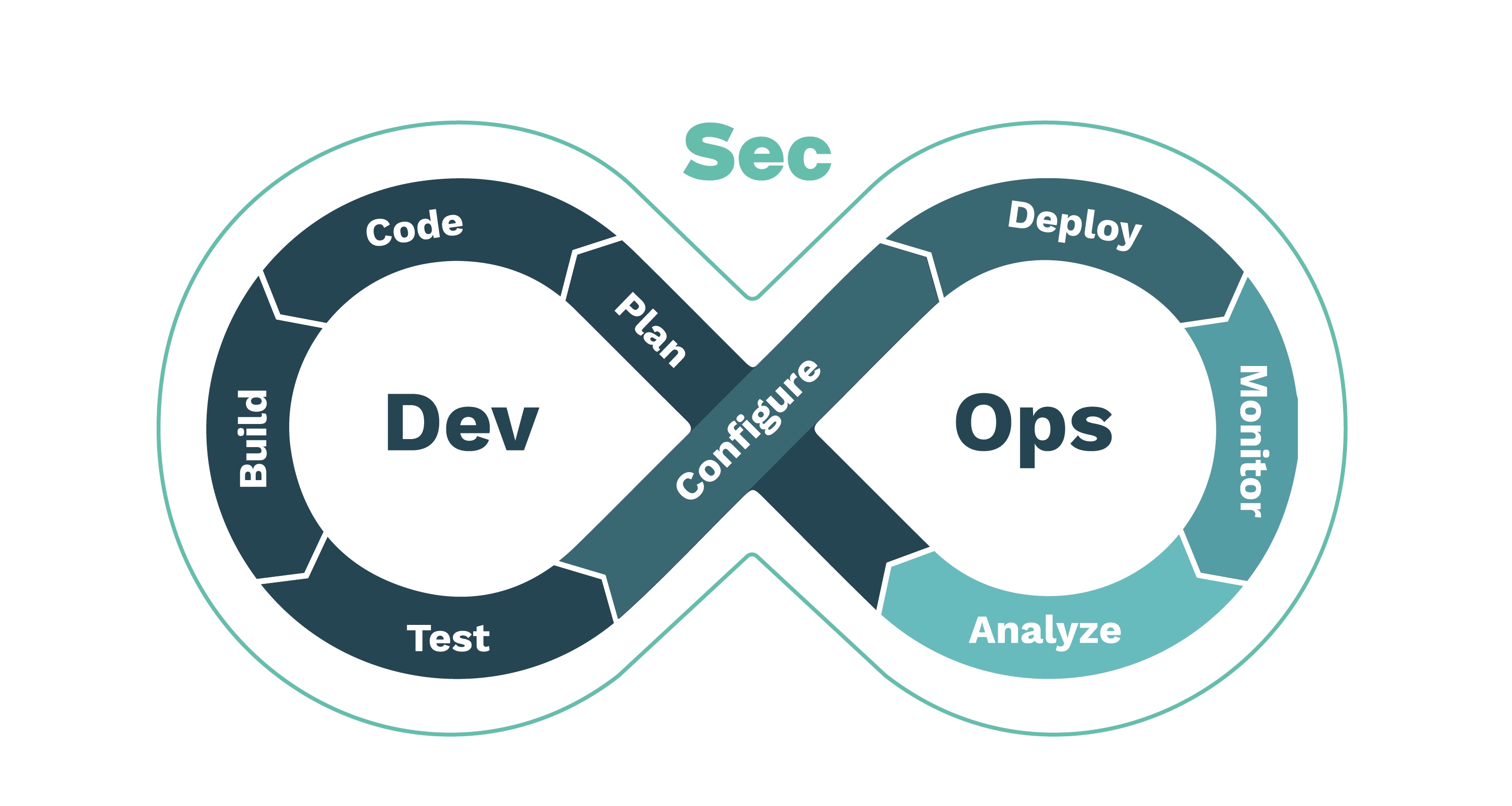}
    \centering
    \caption{A graphical representation of the DevSecOps cycle for robotics.}
    \label{fig:devsecops_cycle}
\end{figure}

\noindent Figure \ref{fig:devsecops_cycle} presents a graphical representation of the DevSecOps cycle in the robotics domain.

 %===============================================================================

%\section{Security, quality and safety}
\section{Discussion and future work}

With DevOps, engineering efforts now serve multiple masters. Where the focus used to be purely on precision, now longevity, scalability and security (with DevSecOps) hold equal footing in product requirement discussions. Where Quality Assurance (QA) had traditionally held the line on risk definition and maintainability, Security now has an equal, if not dominant, role.\\
\newline
\textbf{Quality} (Quality Assurance or QA for short) and \textbf{Security} are often misunderstood when it comes to software. Ivers argues \cite{ivers_2017} that quality "essentially means that the software will execute according to its design and purpose" while "security means that the software will not put data or computing systems at risk of unauthorized access". Within \cite{ivers_2017} one relevant open question that arises is whether the quality problems are also security issues or vice versa. Ivers indicates that quality bugs can turn into security ones, provided they are exploitable, and addresses the question by remarking that quality and security are critical components to a broader notion: software integrity. Coming from the same group, Vamosi \cite{vamosi_2017} argues that "quality code may not always be secure, but secure code must always be quality code". This somehow conflicts with the previous view and leads one to think that secure software is a subset of quality. The authors of this article reject this view and argue instead that Quality and Security remain two separate properties of software that may intersect on certain aspects (e.g. testing).\\
\newline
As nicely indicated by Lopez \cite{lopez_2017} for the DevOps scenario and extending it for security, in the DevSecOps scenario, both security and QA integrate into the testing and development processes and take a collaborative approach. Quality and Security are ensured throughout the testing and delivery cycles and both the testing and development teams are responsible for them. In other words, \emph{compared to the traditional waterfall pattern where quality creeps in toward the end, with DevSecOps security and quality come in at every level}.
\newline
In robotics there is a clear separation between Security and Quality that is best understood with scenarios involving robotic software components. For example, if one was building an industrial Automated Guided Vehicle (AGV), Autonomous Mobile Robots (AMRs) or a self-driving vehicle, often, she/he would need to comply with coding standards (e.g. MISRA \cite{ward2006misra} for developing safety-critical systems).  The same system's communications, however, regardless of its compliance with the coding standards, might rely on a channel that does not provide encryption or authentication and is thereby subject to eavesdropping and man-in-the-middle attacks. Security would be a strong driver here and as remarked by Vamosi \cite{ivers_2017}, "neither security nor quality would be mutually exclusive, there will be elements of both".\\
\newline
Quality in robotics, still on its early stages \cite{pichler2019can} (though much more developed than security in the author's opinion), is often viewed as a pre-condition for \textbf{Safety}-critical systems. Similarly, as argued by several, safety can not be guaranteed without security \cite{goertzel2009software, bagnara2017misra}. Coding standards such as MISRA C have been extended \cite{misra2016amendment1, misra2016addendum2} to become the "C coding standard" of choice for the automotive industry and for all industries developing embedded systems that are safety-critical and/or security-critical \cite{bagnara2017misra}. As introduced by ISO/IEC TS 17961:2013 "in practice, security-critical and safety-critical code have the same requirements". This statement is somehow supported by Goertzel \cite{goertzel2009software} but emphasized the importance of software remaining dependable under extraordinary conditions and the interconnection between safety and security in software. This same argument was later extended by Bagnara \cite{bagnara2017misra} who acknowledges that having embedded systems, non-isolated anymore, plays a key role in the relationship between safety and security. According to Bagnara, "while safety and security are distinct concepts, when it comes to connected software" (non-isolated)  "not having one implies not having the other", referring to integrity.\\
\newline
Acknowledging that both Security and QA are embed into the DevSecOps cycle brings us to question whether safety --connected to Quality-- could also fit in. Myklebust et al. \cite{myklebust2019safety} argue that DevOps, with its frequent changes, make systems’ maintainability – e.g., change impact analysis – a more challenging topic, which makes it more difficult to integrate safety within the DevOps cycle. They do acknowledge though that DevOps is also a considerable trend for non-critical systems and make an interesting connection between safety and security, which indirectly leads to DevSecOps. In a similar line of thought, Johnson \cite{johnson_2018_performance} indicates that the time and cost involved in this process makes the DevOps paradigm of continuous release challenging for safety-critical software, especially in delivering to the end customer. He, however, points out that despite this, "aviation and defense teams have embraced the principles of DevOps, especially at the pre-certification stage, because of the potential for the higher overall quality it offers".\\
\newline
Altogether this leads us to the following questions: how do we apply safety-critical practices to the DevSecOps cycle in robotics? And more importantly, if safety and security coding (software) standards do not guarantee that the final robotic system will be secure, will it be safe given the safety and security connection? Some preliminary work \cite{meyers2019model, di2016software} tackles the first question and indicates that Model-Driven Engineering (MDE) frameworks might allow a mix of DevSecOps with the functional safety process. Further research on these questions remains open for future efforts.

%Altogether this leads us to the following question, which we leave unanswered for future research efforts: How do we apply common QA and safety practices in robotics given the DevSecOps cycle above? If safety and security coding (software) standards do not guarantee that the final robotic system will be secure, will it be safe given safety's connection with security?

 %===============================================================================

\section*{Acknowledgments}
We thank Lander Usategui San Juan and Mike Karamousadakis for their feedback and useful discussions. 
This work has partially been funded by the ROS-Industrial Quality-Assured Robot Software Components (ROSin) RedROS-I and RedROS2-I FTPs which received funding from the European Union’s Horizon 2020 research and innovation programe under the project ROSIN with the Grant Agreement No 732287. This research was also financially supported by the Spanish Government through CDTI Neotec actions (SNEO-20181238).
Special thanks to the Basque Cyber Security Centre BCSC (Basque Government's agency SPRI) for the support in actions fostering awareness in robot Cyber Security. Last but not least, authors are grateful to the local administration Diputación Foral de Álava for the support to entrepreneurship in innovation actions (EMPREM-2019/00002). 

%Special thanks to BIC Araba and the Basque Cybersecurity Centre (BCSC) for the support provided. 
%\section{Outline}

\appendix

\bibliographystyle{IEEEtran}
\bibliography{bibliography}

\begin{thebibliography}{10}
\providecommand{\url}[1]{#1}
\csname url@samestyle\endcsname
\providecommand{\newblock}{\relax}
\providecommand{\bibinfo}[2]{#2}
\providecommand{\BIBentrySTDinterwordspacing}{\spaceskip=0pt\relax}
\providecommand{\BIBentryALTinterwordstretchfactor}{4}
\providecommand{\BIBentryALTinterwordspacing}{\spaceskip=\fontdimen2\font plus
\BIBentryALTinterwordstretchfactor\fontdimen3\font minus
  \fontdimen4\font\relax}
\providecommand{\BIBforeignlanguage}[2]{{%
\expandafter\ifx\csname l@#1\endcsname\relax
\typeout{** WARNING: IEEEtran.bst: No hyphenation pattern has been}%
\typeout{** loaded for the language `#1'. Using the pattern for}%
\typeout{** the default language instead.}%
\else
\language=\csname l@#1\endcsname
\fi
#2}}
\providecommand{\BIBdecl}{\relax}
\BIBdecl

\bibitem{ebert2016devops}
C.~Ebert, G.~Gallardo, J.~Hernantes, and N.~Serrano, ``Devops,'' \emph{Ieee
  Software}, vol.~33, no.~3, pp. 94--100, 2016.

\bibitem{devopsorigin1}
S.~Mezak, ``The origins of devops: What’s in a name?''
  https://devops.com/the-origins-of-devops-whats-in-a-name/, 2018, accessed:
  2020-03-19.

\bibitem{devopsorigin2}
B.~Team, ``The origin of devops,''
  https://bugwolf.com/blog/the-origin-of-devops, 2016, accessed: 2020-03-19.

\bibitem{roche2013adopting}
J.~Roche, ``Adopting devops practices in quality assurance,''
  \emph{Communications of the ACM}, vol.~56, no.~11, pp. 38--43, 2013.

\bibitem{ncsc_uk}
\BIBentryALTinterwordspacing
U.~K. The National Cyber Security~Centre, ``Ncsc guidance for secure
  development and deployment,'' September 2017. [Online]. Available:
  \url{https://github.com/ukncsc/secure-development-and-deployment}
\BIBentrySTDinterwordspacing

\bibitem{lwakatare2016towards}
L.~E. Lwakatare, T.~Karvonen, T.~Sauvola, P.~Kuvaja, H.~H. Olsson, J.~Bosch,
  and M.~Oivo, ``Towards devops in the embedded systems domain: Why is it so
  hard?'' in \emph{2016 49th hawaii international conference on system sciences
  (hicss)}.\hskip 1em plus 0.5em minus 0.4em\relax IEEE, 2016, pp. 5437--5446.

\bibitem{graff2003secure}
M.~Graff and K.~R. Van~Wyk, \emph{Secure coding: principles and
  practices}.\hskip 1em plus 0.5em minus 0.4em\relax " O'Reilly Media, Inc.",
  2003.

\bibitem{seacord2008cert}
R.~C. Seacord, \emph{The CERT C secure coding standard}.\hskip 1em plus 0.5em
  minus 0.4em\relax Pearson Education, 2008.

\bibitem{jones2004secure}
R.~L. Jones and A.~Rastogi, ``Secure coding: building security into the
  software development life cycle,'' \emph{Information Systems Security},
  vol.~13, no.~5, pp. 29--39, 2004.

\bibitem{seacord2005secure}
R.~C. Seacord, \emph{Secure Coding in C and C++}.\hskip 1em plus 0.5em minus
  0.4em\relax Pearson Education, 2005.

\bibitem{bird2016devopssec}
J.~Bird, ``Devopssec: Securing software through continuous delivery,'' 2016.

\bibitem{mohan2016secdevops}
V.~Mohan and L.~B. Othmane, ``Secdevops: Is it a marketing buzzword?-mapping
  research on security in devops,'' in \emph{2016 11th International Conference
  on Availability, Reliability and Security (ARES)}.\hskip 1em plus 0.5em minus
  0.4em\relax IEEE, 2016, pp. 542--547.

\bibitem{apexaisafealgs}
C.~Ho, ``Building safe algorithms in the open, part 1 - design,''
  https://www.apex.ai/post/building-safe-algorithms-in-the-open-part-1-design,
  2020, accessed: 2020-03-19.

\bibitem{basetposter}
A.~Z. Baset and T.~Denning, ``Poster: Ide plugins for secure coding.''

\bibitem{bosu2013peer}
A.~Bosu and J.~C. Carver, ``Peer code review to prevent security
  vulnerabilities: An empirical evaluation,'' in \emph{2013 IEEE Seventh
  International Conference on Software Security and Reliability
  Companion}.\hskip 1em plus 0.5em minus 0.4em\relax IEEE, 2013, pp. 229--230.

\bibitem{bosu2014characteristics}
A.~Bosu, ``Characteristics of the vulnerable code changes identified through
  peer code review,'' in \emph{Companion Proceedings of the 36th International
  Conference on Software Engineering}, 2014, pp. 736--738.

\bibitem{vilches2019introducing}
V.~Mayoral-Vilches, L.~U.~S. Juan, B.~Dieber, U.~A. Carbajo, and
  E.~Gil-Uriarte, ``Introducing the robot vulnerability database (rvd),''
  \emph{arXiv preprint arXiv:1912.11299}, 2019.

\bibitem{barstad2014predicting}
V.~Barstad, M.~Goodwin, and T.~Gj{\o}s{\ae}ter, ``Predicting source code
  quality with static analysis and machine learning.'' in \emph{NIK}, 2014.

\bibitem{russell2018automated}
R.~Russell, L.~Kim, L.~Hamilton, T.~Lazovich, J.~Harer, O.~Ozdemir,
  P.~Ellingwood, and M.~McConley, ``Automated vulnerability detection in source
  code using deep representation learning,'' in \emph{2018 17th IEEE
  International Conference on Machine Learning and Applications (ICMLA)}.\hskip
  1em plus 0.5em minus 0.4em\relax IEEE, 2018, pp. 757--762.

\bibitem{gupta2018intelligent}
A.~Gupta and N.~Sundaresan, ``Intelligent code reviews using deep learning,''
  2018.

\bibitem{pradel2018deepbugs}
M.~Pradel and K.~Sen, ``Deepbugs: A learning approach to name-based bug
  detection,'' \emph{Proceedings of the ACM on Programming Languages}, vol.~2,
  no. OOPSLA, pp. 1--25, 2018.

\bibitem{RSF}
V.~{Mayoral Vilches}, L.~{Alzola Kirschgens}, A.~{Bilbao Calvo}, A.~{Hernández
  Cordero}, R.~{Izquierdo Pisón}, D.~{Mayoral Vilches}, A.~{Muñiz Rosas},
  G.~{Olalde Mendia}, L.~{Usategi San Juan}, I.~{Zamalloa Ugarte},
  E.~{Gil-Uriarte}, E.~{Tews}, and A.~{Peter}, ``Introducing the robot security
  framework (rsf), a standardized methodology to perform security assessments
  in robotics,'' \emph{ArXiv e-prints}, Jun. 2018.

\bibitem{foreman2019vulnerability}
P.~Foreman, \emph{Vulnerability management}.\hskip 1em plus 0.5em minus
  0.4em\relax CRC Press, 2019.

\bibitem{mell2006common}
P.~Mell, K.~Scarfone, and S.~Romanosky, ``Common vulnerability scoring
  system,'' \emph{IEEE Security \& Privacy}, vol.~4, no.~6, pp. 85--89, 2006.

\bibitem{vilches2018towards}
V.~M. Vilches, E.~Gil-Uriarte, I.~Z. Ugarte, G.~O. Mendia, R.~I. Pis{\'o}n,
  L.~A. Kirschgens, A.~B. Calvo, A.~H. Cordero, L.~Apa, and C.~Cerrudo,
  ``Towards an open standard for assessing the severity of robot security
  vulnerabilities, the robot vulnerability scoring system (rvss),'' \emph{arXiv
  preprint arXiv:1807.10357}, 2018.

\bibitem{ivers_2017}
\BIBentryALTinterwordspacing
J.~Ivers, ``Security vs. quality: What's the difference?'' Mar 2017. [Online].
  Available:
  \url{https://www.securityweek.com/security-vs-quality-what’s-difference}
\BIBentrySTDinterwordspacing

\bibitem{vamosi_2017}
\BIBentryALTinterwordspacing
R.~Vamosi, ``Does software quality equal software security?: Synopsys,'' Mar
  2017. [Online]. Available:
  \url{https://www.synopsys.com/blogs/software-security/does-software-quality-equal-software-security/}
\BIBentrySTDinterwordspacing

\bibitem{lopez_2017}
\BIBentryALTinterwordspacing
K.~Lopez, ``The role of qa in the devops world,'' July 2017. [Online].
  Available: \url{https://devops.com/role-qa-devops-world/}
\BIBentrySTDinterwordspacing

\bibitem{ward2006misra}
D.~D. Ward, ``Misra standards for automotive software,'' 2006.

\bibitem{pichler2019can}
M.~Pichler, B.~Dieber, and M.~Pinzger, ``Can i depend on you? mapping the
  dependency and quality landscape of ros packages,'' in \emph{2019 Third IEEE
  International Conference on Robotic Computing (IRC)}.\hskip 1em plus 0.5em
  minus 0.4em\relax IEEE, 2019, pp. 78--85.

\bibitem{goertzel2009software}
K.~M. Goertzel and L.~Feldman, ``Software survivability: where safety and
  security converge,'' in \emph{AIAA Infotech@ Aerospace Conference and AIAA
  Unmanned... Unlimited Conference}, 2009, p. 1922.

\bibitem{bagnara2017misra}
R.~Bagnara, ``Misra c, for security's sake!'' \emph{arXiv preprint
  arXiv:1705.03517}, 2017.

\bibitem{misra2016amendment1}
MISRA, ``Misra c:2012 amendment 1:“additional security guidelines for misra
  c: 2012,”,'' HORIBA MIRA Limited, Nuneaton, Warwickshire, UK, April, Tech.
  Rep., 2016.

\bibitem{misra2016addendum2}
------, ``Misra c:2012 addendum 2 — coverage of misra c:2012 against iso/iec
  ts 17961:2013 “c secure”.'' HORIBA MIRA Limited, Nuneaton, Warwickshire,
  UK, April, Tech. Rep., 2016.

\bibitem{myklebust2019safety}
T.~Myklebust, T.~St{\aa}lhane, and G.~K. Hanssen, ``Safety case and devops
  approach for autonomous cars and ships,'' in \emph{First International
  Workshop on Autonomous Systems Safety}, 2019, p.~95.

\bibitem{johnson_2018_performance}
\BIBentryALTinterwordspacing
M.~Johnson, ``Devops in a safety-critical market,'' September 2018. [Online].
  Available: \url{https://www.psware.com/devops-in-a-safety-critical-market/}
\BIBentrySTDinterwordspacing

\bibitem{meyers2019model}
B.~Meyers, K.~Gadeyne, B.~Oakes, M.~Bernaerts, H.~Vangheluwe, and J.~Denil, ``A
  model-driven engineering framework to support the functional safety
  process,'' in \emph{2019 ACM/IEEE 22nd International Conference on Model
  Driven Engineering Languages and Systems Companion (MODELS-C)}.\hskip 1em
  plus 0.5em minus 0.4em\relax IEEE, 2019, pp. 619--623.

\bibitem{di2016software}
E.~Di~Nitto, P.~Jamshidi, M.~Guerriero, I.~Spais, and D.~A. Tamburri, ``A
  software architecture framework for quality-aware devops,'' in
  \emph{Proceedings of the 2nd International Workshop on Quality-Aware DevOps},
  2016, pp. 12--17.

\end{thebibliography}
\end{document}